%% file: arxivmain.tex
\documentclass{article}

\usepackage{PRIMEarxiv}

\input{arxivpreamble}

\usepackage[utf8]{inputenc}
\usepackage[T1]{fontenc}

\usepackage{graphicx}
\usepackage{booktabs}
\usepackage{caption}
\usepackage{tabularx}
\usepackage{float}
\usepackage{subcaption}
\usepackage{appendix}
\usepackage{amsmath, amssymb, amsfonts}
\usepackage[dvipsnames]{xcolor}
\usepackage{adjustbox}
\usepackage{threeparttable}

\usepackage[colorlinks=true,allcolors=MidnightBlue]{hyperref}
\usepackage{url}

\definecolor{change}{named}{purple}

\title{Latent Dirichlet Transformer VAE for Hyperspectral Unmixing\\with Bundled Endmembers}

\author{
  Giancarlo Giannetti \\
  Ontario Tech University, Canada\\
  \texttt{giancarlo.giannetti@ontariotechu.net}
  \And
  Faisal Z. Qureshi \\
  Ontario Tech University, Canada\\
  \texttt{faisal.qureshi@ontariotechu.ca}
}

\begin{document}

\title{Latent Dirichlet Transformer VAE for Hyperspectral Unmixing\\with Bundled Endmembers}

\date{January 2025}

\maketitle
\begin{abstract}
Hyperspectral images capture rich spectral information that enables per-pixel material identification; however, spectral mixing often obscures pure material signatures. To address this challenge, we propose the Latent Dirichlet Transformer Variational Autoencoder (LDVAE-T) for hyperspectral unmixing. Our model combines the global context modeling capabilities of transformer architectures with physically meaningful constraints imposed by a Dirichlet prior in the latent space. This prior naturally enforces the sum-to-one and non-negativity conditions essential for abundance estimation, thereby improving the quality of predicted mixing ratios. A key contribution of LDVAE-T is its treatment of materials as bundled endmembers, rather than relying on fixed ground truth spectra. In the proposed method our decoder predicts, for each endmember and each patch, a mean spectrum together with a structured (segmentwise) covariance that captures correlated spectral variability. Reconstructions are formed by mixing these learned bundles with Dirichlet-distributed abundances garnered from a transformer encoder, allowing the model to represent intrinsic material variability while preserving physical interpretability. We evaluate our approach on three benchmark datasets, Samson, Jasper Ridge, and HYDICE Urban and show that LDVAE-T consistently outperforms state-of-the-art models in abundance estimation and endmember extraction, as measured by root mean squared error and spectral angle distance, respectively. 
\end{abstract}



\section{Introduction}

Hyperspectral images capture data across a wide range of wavelengths, typically over 100, spanning the infrared, visible, and sometimes ultraviolet spectra. Each material within a scene exhibits a unique spectral signature, enabling per-pixel material identification through appropriate data processing. In satellite-based remote sensing, however, each pixel often corresponds to a large ground area containing multiple materials. This results in spectral mixing, which necessitates hyperspectral unmixing, the process of decomposing mixed pixel spectra into their constituent endmembers.

\begin{figure*}[t]
\centerline{
\includegraphics[width=\linewidth]{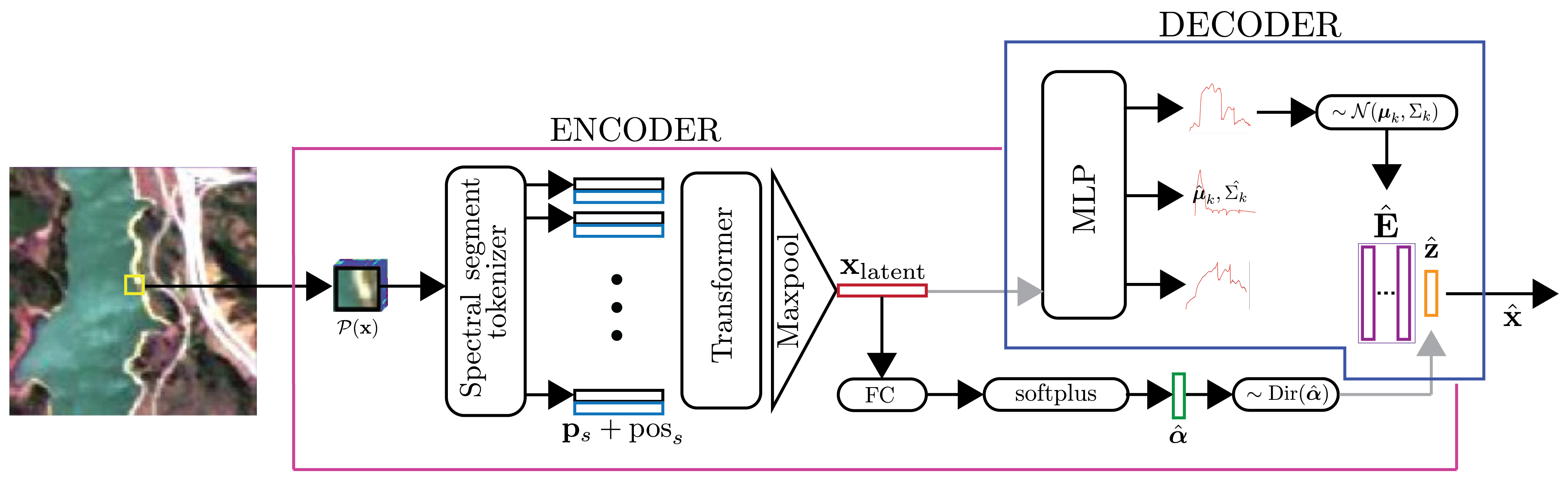}
}
\caption{Overview.  Transformer based encoder takes a hyperspectral image patch $\mathcal{P}(\mathbf{x})$ and construct $\mathbf{x}_\text{latent}$, $\hat{\boldsymbol{\alpha}}$ and abundances $\hat{\mathbf{z}}$.  The decoder takes $\mathbf{x}_\text{latent}$ and $\hat{\mathbf{z}}$ and reconstructs pixel $\mathbf{x}$.  The decoder also constructs endmembers $\hat{\mathbf{e}}_k$.}
\label{fig:arch}
\end{figure*}

Hyperspectral Unmixing (HU) involves decomposing a hyperspectral image into its constituent materials, referred to as endmembers, and estimating the relative abundance of each material within every pixel. This task is essential in many remote sensing applications, as each pixel often represents a mixture of materials due to limited spatial resolution. Traditional HU methods are broadly categorized into linear and nonlinear models. Linear mixing models (LMMs) assume that the observed spectrum is a convex combination of endmember signatures, weighted by their abundance fractions. While computationally efficient and physically interpretable, LMMs often fail to capture complex light interactions such as scattering and nonlinearity, especially in densely vegetated or urban areas.

Nonlinear unmixing models aim to address these limitations by incorporating physical or data-driven nonlinearities into the mixing process. However, these approaches typically involve greater model complexity and higher computational cost. More recently, machine learning and deep learning methods have emerged as powerful tools for HU, offering improved flexibility in modeling spectral and spatial patterns. These include autoencoders, variational models, graph-based networks, and transformer architectures, many of which incorporate priors or constraints (e.g., sparsity, non-negativity, sum-to-one) to enhance physical interpretability. Despite these advances, challenges remain in generalizing across diverse scenes, extracting pure endmembers from limited ground truth, and balancing model expressiveness with physical realism. As such, HU continues to be an active area of research at the intersection of remote sensing, signal processing, and machine learning.

State-of-the-art (SOTA) methods increasingly rely on Vision Transformer (ViT) architectures~\cite{16x16words,DuanUnDat,GhoshTransformer,YangTransformer}, which outperform traditional multilayer perceptron (MLP) and convolutional neural network (CNN) approaches by effectively capturing long-range dependencies~\cite{16x16words}, a critical capability for estimating abundances in hyperspectral data. In contrast, CNNs are inherently limited to local neighborhood information, which may be insufficient for modeling complex spectral-spatial interactions.

To better address the unique characteristics of hyperspectral images, such as locally homogeneous regions, nonlinear mixing effects, and limited labeled data—researchers have proposed several specialized modules to enhance unmixing performance. One promising direction involves incorporating Dirichlet distributions into a model’s latent space~\cite{kiran,soham}, enabling Variational Autoencoders (VAEs) to better model the constraints of hyperspectral unmixing. However, such Dirichlet-based approaches have thus far been limited to MLP and CNN backbones, and have not yet been explored in transformer-based architectures. To bridge this gap, we propose LDVAE-T, which integrates Dirichlet priors into a ViT-based VAE framework, combining the probabilistic interpretability of Dirichlet modeling with the global feature extraction capabilities of transformers. Our key novelty is a “bundled endmember” decoder: instead of using fixed spectra, the model predicts, for each endmember and for each patch, a distributional prototype composed of a mean spectrum and a structured (segmentwise) covariance that captures correlated spectral variability. Reconstructions are then formed by mixing samples form these learned bundles according to Dirichlet-distributed abundances. Further, we tokenize spectra as fixed-length segments to respect spectral locality while enabling efficient attention. Together, these design choices allow LDVAE-T to model intrinsic material variability while preserving the sum-to-one and non-negativity properties central to HU, leading to improved endmember extraction and abundance estimation across diverse scenes.

\section{Related Work}

Pixel unmixing approaches in hyperspectral imaging can be broadly categorized into two groups: (a) physics-based methods and (b) data-driven techniques. Physics-based methods rely on models that describe how light interacts with materials, such as Hapke’s Bidirectional Reflectance Distribution Function (BRDF)\cite{phy1} and the Atmospheric Dispersion Model~\cite{phy2}. While these approaches offer physically grounded interpretations, they often require detailed, scene-specific radiative parameters, limiting their practicality in real-world applications. In contrast, data-driven methods dominate the field due to their flexibility and ease of deployment, though they are highly dependent on the availability of annotated training data. Hybrid models that fuse physics-based insights with data-driven learning have also been studied~\cite{phy3}.  A widely adopted class of data-driven methods is based on Non-Negative Matrix Factorization (NMF), which models the hyperspectral image as the product of two matrices: one representing endmembers and the other representing abundances.  Several extensions to NMF based methods have been proposed in the literature that exploit spatial or spectral neighbourhood structure~\cite{nnmf1,nnmf2,nnmf4}, use iterative refinement to improve results~\cite{nnmf3}, or integrate handcrafted and learned priors for improved generalization~\cite{NMFQMV}.


Recent efforts have shifted toward deep learning-based approaches, which offer strong performance in modeling spectral-spatial dependencies. DeepGUn~\cite{DeepGUn} combines deep latent representation learning with vertex component analysis (VCA) for endmember extraction. Autoencoder-based models have also gained traction, with CNNAEU~\cite{CNNAEU} being the first to introduce a convolutional autoencoder for hyperspectral unmixing. The LDVAE model~\cite{kiran} further advances this line by incorporating a latent Dirichlet distribution within a variational autoencoder (VAE) framework. The Dirichlet prior naturally satisfies the non-negativity and sum-to-one constraints of the unmixing problem while enabling a probabilistic interpretation of abundances. Building on this, SpACNN-LDVAE~\cite{soham} replaces the MLP encoder with a CNN encoder, demonstrating improved performance by preserving spatial structure during training. Ghosh \etal introduced the first ViT-based model for hyperspectral unmixing, DeepTrans-HsU~\cite{GhoshTransformer}, which showed that vision transformers can outperform previous deep learning approaches with minimal architectural changes. UnDAT~\cite{DuanUnDat} employs a transformer encoder-decoder backbone along with spectral and spatial clustering modules to enhance unmixing performance, albeit at the cost of higher computational complexity. SSF-Net~\cite{SSFNet}, an autoencoder-based model, introduces a spatial-spectral fusion module designed to capture local spatial heterogeneity and spectral diversity among endmembers more effectively.

\begin{figure}
  \centering
  \includegraphics[width=1.02\linewidth]{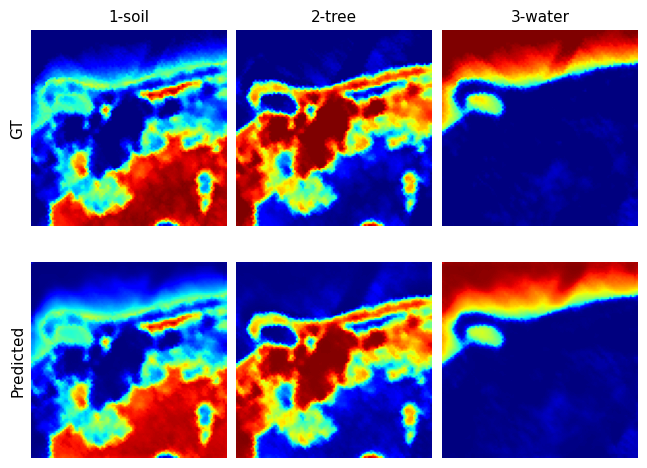}
  \caption{Ground Truth vs Predicted endmember heat-maps for Samson dataset.  These plots visualize per-pixel abundances.}
  \label{fig:samson_heatmaps}
\end{figure}

\begin{figure}
  \centering
  \includegraphics[width=1.02\linewidth]{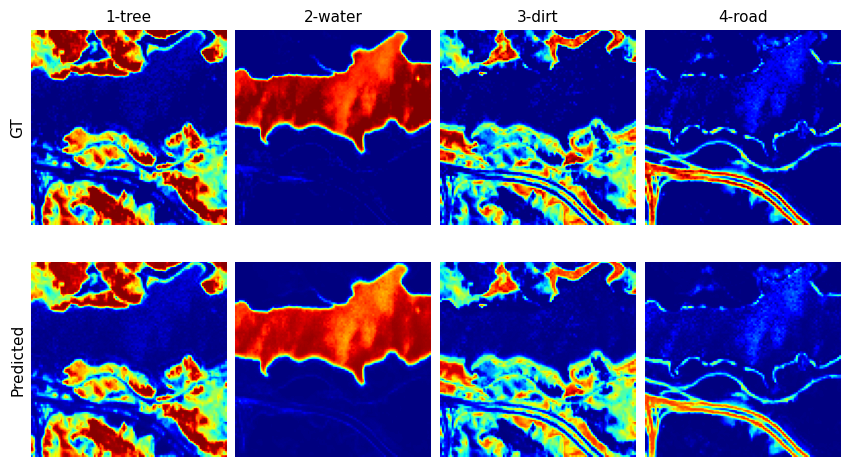}
  \caption{Ground Truth vs. Predicted endmember heat-maps for Jasper Ridge dataset.  These plots visualize per-pixel abundances.}
  \label{fig:jasper_heatmaps}
\end{figure}

\begin{figure*}
  \centering
  \includegraphics[width=1.02\linewidth]{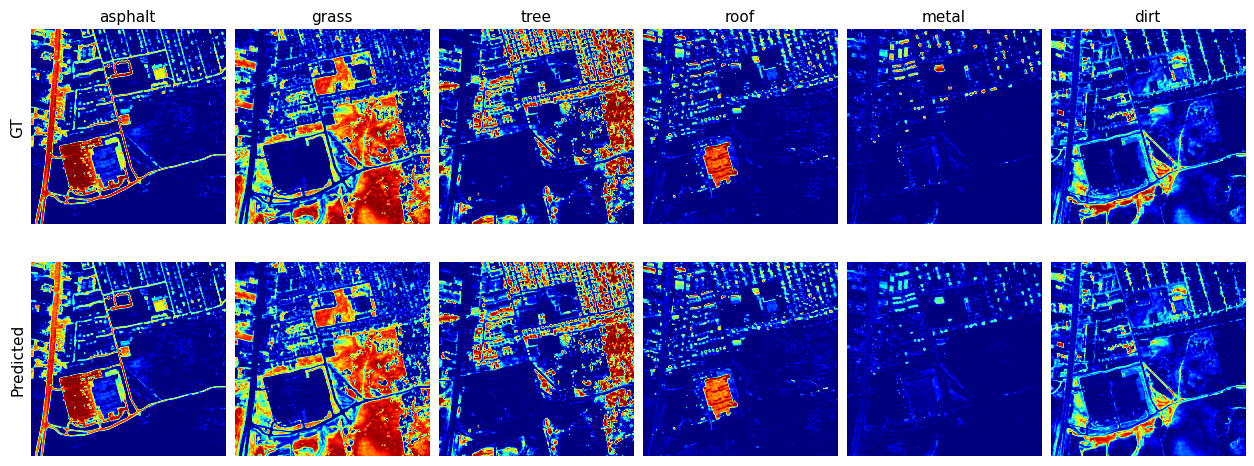}
  \caption{Ground Truth vs Predicted endmember heat-maps for HYDICE Urban dataset.  These plots visualize per-pixel abundances.}
  \label{fig:hydice_heatmaps}
\end{figure*}

\section{Method}

\newcommand{\be}{\mathbf{e}}
\newcommand{\bx}{\mathbf{x}}
\newcommand{\bE}{\mathbf{E}}
\newcommand{\ba}{\mathbf{a}}
\newcommand{\bn}{\mathbf{n}}
\newcommand{\bz}{\mathbf{z}}
\newcommand{\bp}{\mathbf{p}}
\newcommand{\bh}{\mathbf{h}}
\newcommand{\balpha}{\boldsymbol{\alpha}}
\newcommand{\W}{\mathbf{W}}

In hyperspectral imaging, we often represent each pixel $\bx \in \mathbb{R}^C$ using
\begin{equation*}
  \bx = \bE \bz + \bn
\end{equation*}
where $C$ refers to the number of channels, $\bE \in \mathbb{R}^{C \times K}$ is the endmember matrix that collects $K$ endmembers column-wise, $\bz \in \mathbb{R}^{K \times 1}$ is the abundance vector  representing the proportion of each endmember in pixel $\bx$, and $\bn$ is additive noise. Here the abundance vector follows non-negative and sum-to-one constraints, i.e., $\forall_{k \in K} a_k \ge 0$ and $\sum_{k \in K} z_k = 1$.  Unmixing computes abundances $\ba$ and endmembers matrix $\bE$ given a hyperspectral pixel $\bx$.  A straightforward extension is to consider a local patch $\mathcal{N}(\bx)$ around $\bx$ when performing unmixing.

\subsection{LDVAE-T Model}

\begin{figure*}[ht!]
  \centering
  \includegraphics[width=.96\textwidth]{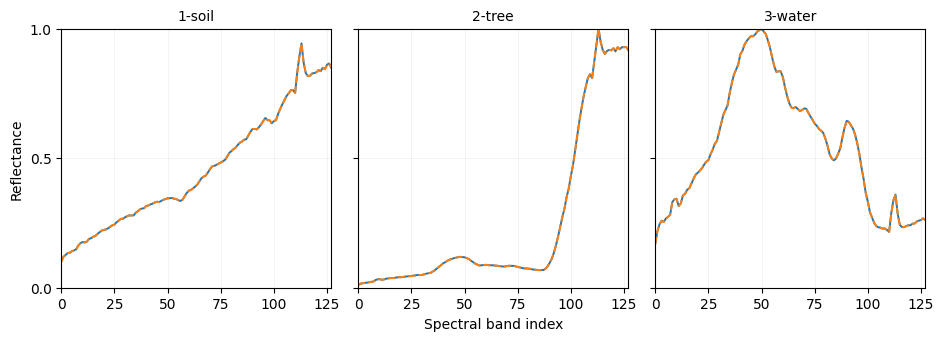}
  \caption{Extracted (orange) vs ground truth (blue) spectral signatures for endmembers in Samson dataset.}
  \label{fig:SamsonSpectra}
\end{figure*}

\begin{table*}[ht!]
\renewcommand{\arraystretch}{1.3}
\caption{SAD Scores for Endmember Extraction on Samson Dataset}
\label{tab:SAD_Samson}
\centering{
\fontsize{12.5}{15.5}\selectfont
\begin{adjustbox}{max width=.98\textwidth}
\begin{tabular}{@{}lcccccccc@{}}
\toprule
   & NMF-QMV~\cite{NMFQMV} & CNNAEU~\cite{CNNAEU}  & LDVAE~\cite{kiran}  & SpACNN-LDVAE~\cite{soham} & DeepTrans-HsU~\cite{GhoshTransformer}  &  UnDAT~\cite{DuanUnDat}  & SSF-Net~\cite{SSFNet} & LDVAE-T \\ 
Endmember&2022&2021&2024&2024&2022&2023&2024&2025\\\midrule
soil        & 0.02326 & 0.07565 & 0.0959 & 0.2097 & 0.0128 & 0.01191 & 0.0092 & \textbf{0.000114}~\textcolor{OliveGreen}{\scriptsize$\uparrow$98.76\%} \\
tree        & 0.06086 & 0.05440 & 1.2788 & 0.5347 & 0.0674 & 0.03775 & 0.0314 & \textbf{0.000202}~\textcolor{OliveGreen}{\scriptsize$\uparrow$99.36\%} \\
water       & 1.45643 & 0.03642 & 0.4022 & 0.8233 & 0.0729 & 0.00813 & 0.0373 & \textbf{0.000213}~\textcolor{OliveGreen}{\scriptsize$\uparrow$97.38\%} \\ \midrule
average     & 0.51352 & 0.05549 & 0.5923 & 0.5525 & 0.0510 & 0.01926 & 0.0260 & \textbf{0.000177}~\textcolor{OliveGreen}{\scriptsize$\uparrow$99.08\%} \\ \bottomrule
\end{tabular}
\end{adjustbox}
}
\end{table*}

\begin{table*}[ht!]
\renewcommand{\arraystretch}{1.3}
\caption{RMSE Scores for Abundance Estimation on Samson Dataset}
\label{tab:RMSE_Samson}
\centering{
\fontsize{12.5}{15.5}\selectfont
\begin{adjustbox}{max width=.98\textwidth}
\begin{tabular}{lcccccccc}
\toprule
   & NMF-QMV~\cite{NMFQMV} & CNNAEU~\cite{CNNAEU}  & LDVAE~\cite{kiran}  & SpACNN-LDVAE~\cite{soham} & DeepTrans-HsU~\cite{GhoshTransformer}  &  UnDAT~\cite{DuanUnDat}  & SSF-Net~\cite{SSFNet} & LDVAE-T \\ 
Endmember&2022&2021&2024&2024&2022&2023&2024&2025\\\midrule
soil        & 0.23298 & 0.3157  & 0.2609 & 0.2097 & 0.0712 & 0.04306 & 0.0511  & \textbf{0.03032}~\textcolor{OliveGreen}{\scriptsize$\uparrow$29.59\%} \\
tree        & 0.24432 & 0.2911  & 0.3431 & 0.5347 & 0.0683 & 0.02854 & 0.0502  & \textbf{0.02401}~\textcolor{OliveGreen}{\scriptsize$\uparrow$15.87\%} \\
water       & 0.37621 & 0.1552  & 0.3165 & 0.2098 & 0.0930 & 0.03128 & 0.0272  & \textbf{0.02249}~\textcolor{OliveGreen}{\scriptsize$\uparrow$17.32\%} \\ \midrule
average     & 0.28450 & 0.2540  & 0.3078 & 0.2412 & 0.0783 & 0.03429 & 0.0428  & \textbf{0.02561}~\textcolor{OliveGreen}{\scriptsize$\uparrow$25.31\%} \\ \bottomrule
\end{tabular}
\end{adjustbox}
}
\end{table*}

The proposed \emph{LDVAE-T} architecture consists of a transformer-based encoder that process a local hyperspectral image patch $\mathcal{P}(\bx)$ and predicts Dirichlet concentration parameters $\balpha$ (abundance prior) for pixel $\bx$.  Additionally, it emits a latent code for $\bx$ that the decoder uses to reconstruct the pixel's endmember spectra.  Recall that endmember spectra are shared by all pixels.
Under the assumption that each endmember follows a Gaussian distribution, the decoder maps the latent code to the Gaussian parameters---mean $\mu_k$ and covariance $\Sigma_k$---for each endmember $k$.  Next, we sample endmembers and mix them explicitly.  For each endmember $k$, draw $\be_k \sim \mathcal{N}(\boldsymbol{\mu}_k, \Sigma_k)$ and draw abundances $\bz \sim \mathrm{Dir}(\balpha)$.  The pixel is then reconstructed by the mixture $\hat{\bx} = \phi (\sum_k z_k \be_k)$.  

Figure~\ref{fig:arch} illustrates the model. The tokenizer converts the local patch $\mathcal{P}(\bx)$ into $S$ tokens $\bp_s \in \mathbb{R}^d$ for $s \in \{1, \cdots, S \}$.  Learned positional encodings are added to each token,
\[
\tilde{\bp}_s = \bp_s + \mathrm{pos}_s.
\]

The sequence $\{ \tilde{\bp}_s\}_{s=1}^S$ is passed to a transformer encoder with $L=4$ layers; each layer uses multi-head self-attention with $H=16$ heads followed by a position-wise feed-forward network (with residual connections and layer normalization).  Self-attention enables the model to capture long-range dependencies across both spectral bands and spatial locations within the patch, yielding a comprehensive representation of material mixtures.  The encoder outputs $\{ \bh_s \}_{s=1}^S$ are max-pooled to construct the latent code for $\bx$
\[
\bx_{\text{latent}} = \max_{s \in \{1,\cdots,S\}} \bh_s \text{\ (element-wise max)}.
\]
The latent vector is passed through a softplus to produce the Dirichlet concentration parameters
\[
\balpha = \text{softplus}( \W \bx_{\text{latent}} + \mathbf{b}) + \epsilon \mathbf{1},\ \ \ \balpha \in \mathbb{R}^K_{>0}.
\]
Here $\W$ and $\mathbf{b}$ are affine parameters for the soft-plus layer and small $\epsilon > 0$ ensures strictly positive entries.  Sampling from $\text{Dir}(\balpha)$ yields an abundance vector $\bz$ that satisfies non-negativity and sum-to-one constraints.

The decoder comprises of two MLPs.  First, MLP takes the latent vector $\bx_{\text{latent}}$ and predicts endmembers means $\boldsymbol{\mu}_k$ and covariances $\Sigma_k$.  The second MLP takes abundance vector $\bz$ and sampled endmembers $\be_k \sim \mathcal{N}(\boldsymbol{\mu}_k, \Sigma_k)$ and reconstruct pixel $\hat{\bx}$.  Similar to other approaches, we assume that the number of endmembers $K$ is known \textit{a priori}.

We employ the following losses during training.  \textbf{Abundance loss} penalizes the divergence between predicted and ground truth abundances
\[
\mathcal{L}_{\text{abundance}}= \text{MSE}(\hat{\bz}, \bz)
\]
where $\hat{\bz}$ are the predicted abundances for pixel $\bx$ and $\bz$ are the ground-truth abundances.  

\textbf{Endmember bundles loss} minimizes the KL divergence between the predicted endmember Gaussian and the ground-truth endmember bundle (estimated from \textit{pure pixels}) 
\[
\mathcal{L}_{\text{endmembers}} = \sum_{k=1}^K \mathbf{\alpha}_k\text{KL} \left( \mathcal{N}(\hat{\boldsymbol{\mu}}_k, \hat{\Sigma}_k)\; \big\|\;
\mathcal{N}(\boldsymbol{\mu}_k, \Sigma_k) \right).
\]
Where $\mathbf{\alpha}_k$ is the predicted pre-sampled abundance estimation of endmember $k$. This mixture-weighted regularization constrains the loss calculation to endmembers present in the given pixel.

Lastly, the latent Dirichelet variational autoencoder optimizes
\[
\mathcal{L}_{\text{ELBO}} = \mathbf{E}_{q_\theta(\bz | \bx) } [ \log p_\phi(\bx | \bz) ] - \text{KL}( q_\theta(\bz|\bx)\; \big\|\;  p(\bz)   ).
\]
with reconstruction term
\[
\mathcal{L}_{\text{recon}} = MSE(\hat{\bx}, \bx)
\]
and Dirichlet prior $p(\bz) = \text{Dir}(\balpha^{\text{prior}})$.  When $q_\theta(\bz|\bx) = \text{Dir}(\hat{\balpha})$ 
\begin{align*}
\text{KL}\left( \text{Dir}(\hat{\balpha})\;\big\|\;\text{Dir}(\balpha^{\text{prior}})\right)
  &= \sum \log \Gamma(\alpha_k^{\text{prior}}) \\
  &- \sum \log \Gamma(\hat{\alpha}_k) \\
  &+ \sum (\hat{\alpha}_k - \alpha_k^{\text{prior}}) \frac{d}{dx} \ln \Gamma(\hat{\alpha}_k).
\end{align*}
Here $\hat{\balpha}$ is the concentration parameter of the estimated Dirichlet distribution and 
$\balpha^{\text{prior}}$ is the concentration parameter of the Dirichlet prior. $\Gamma(\cdot)$ is the Gamma function.
Thus the overall loss is
\[
\mathcal{L}=\mathcal{L}_{\text{ELBO}} + \lambda_\text{abundances}  \mathcal{L}_{\text{abundances}} + \lambda_\text{endmembers} \mathcal{L}_{\text{endmembers}}
\]
with $\lambda_\text{abundances}$ and $\lambda_\text{endmembers}$ control the relative strengths of abundances and endmember terms.  We set $\lambda_\text{abundances}$ to $1$, and apply a heavy anneal to $\lambda_\text{endmembers}$ for experiments reported here. The annealing begins at $1\times10^{-6}$ and progresses towards $1.0$ over 80000 epochs (which is never reached). This is done to keep $\mathcal{L}_{\text{endmembers}}$ from significantly outweighing the other terms by a magnitude of $1\times10^{6}$ at the start of training.

\begin{figure*}[ht!]
  \centering
  \includegraphics[width=.96\textwidth]{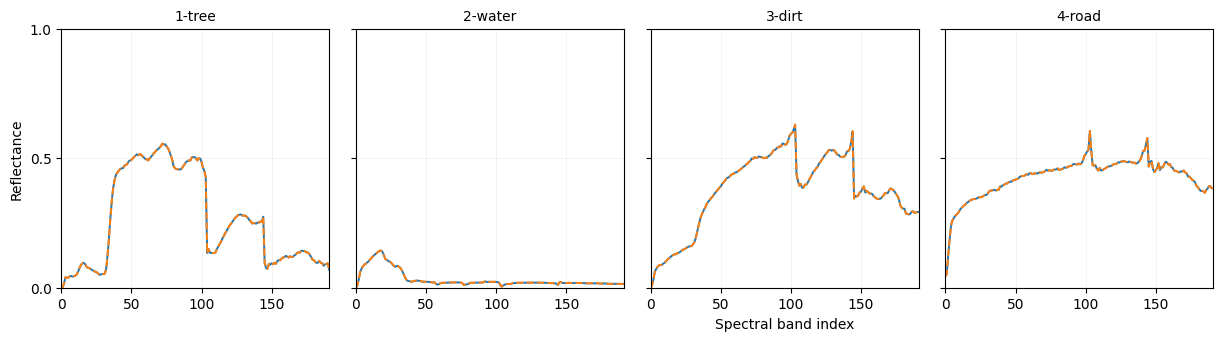}
  \caption{Extracted (orange) vs ground truth (blue) spectral signatures for endmembers in Jasper Ridge dataset.}
  \label{fig:JasperSpectra}
\end{figure*}

\begin{table*}[ht!]
\renewcommand{\arraystretch}{1.3}
\caption{SAD Scores for Endmember Extraction on Jasper Ridge Dataset}
\label{tab:SAD_Jasper}
\centering{
\fontsize{12.5}{15.5}\selectfont
\begin{adjustbox}{max width=.98\textwidth}
\begin{tabular}{lcccccccc}
\toprule
   & NMF-QMV~\cite{NMFQMV} & CNNAEU~\cite{CNNAEU}  & LDVAE~\cite{kiran}  & SpACNN-LDVAE~\cite{soham} & DeepTrans-HsU~\cite{GhoshTransformer}  &  UnDAT~\cite{DuanUnDat}  & SSF-Net~\cite{SSFNet} & LDVAE-T \\ 
Endmember&2022&2021&2024&2024&2022&2023&2024&2025\\\midrule
tree        & 0.05016 & 0.3104 &   -   &   -   &   -   & 0.04519 & 0.0781 & \textbf{0.000255}~\textcolor{OliveGreen}{\scriptsize$\uparrow$99.44\%} \\
water       & 0.28387 & 0.6082 &   -   &   -   &   -   & 0.02811 & 0.0293 & \textbf{0.001269}~\textcolor{OliveGreen}{\scriptsize$\uparrow$95.49\%} \\
soil        & 0.17326 & 0.3381 &   -   &   -   &   -   & 0.09074 & 0.0484 & \textbf{0.000270}~\textcolor{OliveGreen}{\scriptsize$\uparrow$99.44\%} \\
road        & 1.46974 & 0.0519 &   -   &   -   &   -   & 0.03306 & 0.0238 & \textbf{0.000288}~\textcolor{OliveGreen}{\scriptsize$\uparrow$98.79\%} \\ \midrule
average     & 0.49426 & 0.3271 &   -   &   -   &   -   & 0.04927 & 0.0449 & \textbf{0.000520}~\textcolor{OliveGreen}{\scriptsize$\uparrow$98.84\%} \\ \bottomrule
\end{tabular}
\end{adjustbox}
}
\end{table*}

\begin{table*}[ht!]
\renewcommand{\arraystretch}{1.3}
\caption{RMSE Scores for Abundance Estimation on Jasper Ridge Dataset}
\label{tab:RMSE_Jasper}
\centering{
\fontsize{12.5}{15.5}\selectfont
\begin{adjustbox}{max width=.98\textwidth,}
\begin{tabular}{lcccccccc}
\toprule
   & NMF-QMV~\cite{NMFQMV} & CNNAEU~\cite{CNNAEU}  & LDVAE~\cite{kiran}  & SpACNN-LDVAE~\cite{soham} & DeepTrans-HsU~\cite{GhoshTransformer}  &  UnDAT~\cite{DuanUnDat}  & SSF-Net~\cite{SSFNet} & LDVAE-T \\ 
Endmember&2022&2021&2024&2024&2022&2023&2024&2025\\\midrule
tree        & 0.12528 & 0.3169 &   -   &   -   &   -   & 0.06031 & 0.0721 & \textbf{0.02609}~\textcolor{OliveGreen}{\scriptsize$\uparrow$56.74\%} \\
water       & 0.20375 & 0.2118 &   -   &   -   &   -   & 0.04211 & 0.0761 & \textbf{0.02892}~\textcolor{OliveGreen}{\scriptsize$\uparrow$31.32\%} \\
soil        & 0.14647 & 0.2978 &   -   &   -   &   -   & 0.07007 & 0.0930 & \textbf{0.03855}~\textcolor{OliveGreen}{\scriptsize$\uparrow$44.98\%} \\
road        & 0.18446 & 0.2043 &   -   &   -   &   -   & 0.07792 & 0.0782 & \textbf{0.03819}~\textcolor{OliveGreen}{\scriptsize$\uparrow$50.99\%} \\ \midrule
average     & 0.16499 & 0.2577 &   -   &   -   &   -   & 0.06260 & 0.0798 & \textbf{0.03294}~\textcolor{OliveGreen}{\scriptsize$\uparrow$47.38\%} \\ \bottomrule
\end{tabular}
\end{adjustbox}
}
\end{table*}

\section{Experimental Setup}

We evaluate our model on three widely used hyperspectral unmixing benchmarks: Samson~\cite{samsondataset}, Jasper Ridge~\cite{JasperRidge}, and HYDICE Urban~\cite{hydiceurban}. The Samson dataset contains a $95 \times 95$ hyperspectral image with 156 spectral bands and three ground-truth endmembers: \emph{Soil}, \emph{Tree}, and \emph{Water}. Jasper Ridge comprises a $100 \times 100$ image with 198 spectral bands and four ground-truth endmembers: \emph{Tree}, \emph{Water}, \emph{Dirt}, and \emph{Road}. HYDICE Urban has $307 \times 307$ pixels with 162 spectral bands and is available in three variants containing four, five, or six endmembers. We use the six-endmember variant: \emph{Asphalt Road}, \emph{Grass}, \emph{Tree}, \emph{Roof}, \emph{Metal}, and \emph{Dirt}. All three datasets provide per-pixel ground-truth abundance maps. 

These datasets provide a single ``true'' spectrum for each endmember. In practice, endmember spectra are not unique: the same material can exhibit measurable spectral variability under different illumination, viewing geometry, sensor characteristics, and environmental conditions.  We first identify high-purity pixels using the Pixel Purity Index (PPI)---pixels dominated by a single endmember---and then use their spectra to initialize the spectral distribution of each endmember. We model each endmember’s spectral variability with a (multivariate) Gaussian distribution.  We refer to these as endmember bundles, that is, each endmember is represented by a distribution over spectra rather than a single spectral signature.

We adopt a 20/80 train–test split for training and evaluation.  The model is trained in a supervised setting for 1000 epochs using a batch size of 128 and the Adam optimizer with a learning rate of $2 \times 10^{-4}$. During training, zero-padding is applied to the image boundaries to maintain spatial consistency in patch extraction.

\subsection{Evaluation Metrics}

We capture endmember reconstruction using Spectral Angle Distance (SAD), which quantifies the discrepancy between predicted and ground-truth endmember spectra by measuring the angle between their spectral vectors in a high-dimensional space, making it invariant to per-pixel intensity (scaling) changes:
\[
\text{SAD}(\hat{\be},\be) = \cos^{-1} \left( \frac{\hat{\be}_k^\top \be_k}{\|\hat{\be}_k\| \, \|\be_k\|} \right),
\]
where \( \hat{\be}_k \) is the predicted endmember distribution's mean and \( \be_k \) is the mean spectrum of the corresponding GT endmember bundle.  Lower SAD values indicate closer alignment between the predicted and reference spectra. We use SAD to assess endmember-reconstruction accuracy for each material class in the dataset.

We evaluate abundance-estimation accuracy using the Root Mean Squared Error (RMSE), which measures the average deviation between predicted and ground-truth abundances across pixels:
\[
\text{RMSE}(\hat{\bz},\bz) = \sqrt{ \frac{1}{N} \sum_{n=1}^N \| \hat{\bz}_n - \bz_n \|^2  }
\]
where $\hat{\bz}$ and $\bz$ refers to the predicted and the ground truth abundances.  $N$ is the number of pixels.  A lower RMSE value indicates better abundance estimation.

\begin{figure*}[ht!]
  \centering
  \includegraphics[width=.96\textwidth]{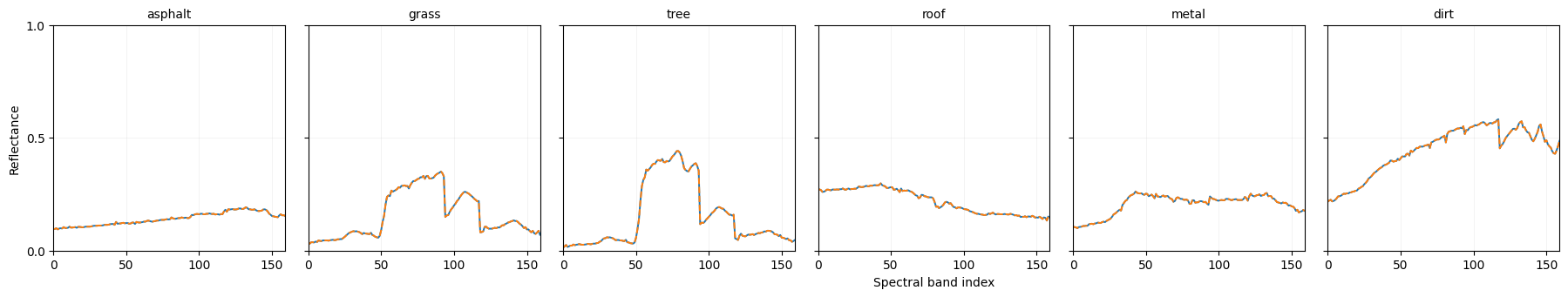}
  \caption{Extracted (orange) vs ground truth (blue) spectral signatures for endmembers in HYDICE Urban dataset.}
  \label{fig:HYDICESpectra}
\end{figure*}

\begin{table*}[ht!]
\renewcommand{\arraystretch}{1.3}
\caption{SAD Scores for Endmember Extraction on HYDICE Urban Dataset.  $^*$SSF-Net used a five endmember version of HYDICE Urban for their experiments.}
\label{tab:SAD_Urban}
\small
\centering{
\begin{adjustbox}{max width=.98\textwidth}
\begin{tabular}{lcccccc}
\toprule
   &SGSNMF~\cite{nnmf2}&SSWNMF~\cite{nnmf1}& LDVAE~\cite{kiran}  & SpACNN-LDVAE~\cite{soham} &SSF-Net$^*$~\cite{SSFNet} & LDVAE-T \\ 
Endmember&2017&2022&2024&2024&2024&2025\\\midrule
Asphalt road    &0.0841&0.0782& 0.4262 & 0.2786 &  0.0629  & \textbf{0.000216}~\textcolor{OliveGreen}{\scriptsize$\uparrow$99.66\%} \\
Grass           &0.1516&0.1490& 0.3323 & 0.1936 &  0.0411  & \textbf{0.000145}~\textcolor{OliveGreen}{\scriptsize$\uparrow$99.65\%} \\
Tree            &0.1199&0.1173& 0.3177 & 0.4411 &  0.0850  & \textbf{0.000155}~\textcolor{OliveGreen}{\scriptsize$\uparrow$99.82\%} \\
Roof            &0.0731&0.0713& 0.4393 & 0.4502 &  0.0487  & \textbf{0.000208}~\textcolor{OliveGreen}{\scriptsize$\uparrow$99.57\%} \\
Metal           &0.1250&0.1241& 0.7004 & 0.3241 &  -       & \textbf{0.000338}~\textcolor{OliveGreen}{\scriptsize$\uparrow$99.73\%} \\
Dirt            &0.0859&0.0802& 0.2806 & 0.2026 &  0.1065  & \textbf{0.000088}~\textcolor{OliveGreen}{\scriptsize$\uparrow$99.89\%} \\ \midrule
Average         &0.1060&0.1034& 0.4161 & 0.3151 &  0.0688  & \textbf{0.000192}~\textcolor{OliveGreen}{\scriptsize$\uparrow$99.72\%} \\ \bottomrule
\end{tabular}
\end{adjustbox}
}
\end{table*}

\begin{table*}[ht!]
\renewcommand{\arraystretch}{1.3}
\caption{RMSE Scores for Abundance Estimation on HYDICE Urban Dataset.  $^*$SSF-Net used a five endmember version of HYDICE Urban for their experiments.}
\label{tab:RMSE_Urban}
\small
\centering{
\begin{adjustbox}{max width=.98\textwidth}
\begin{tabular}{lcccccc}
\toprule
   &SGSNMF~\cite{nnmf2}&SSWNMF~\cite{nnmf1}& LDVAE~\cite{kiran}  & SpACNN-LDVAE~\cite{soham} &SSF-Net$^*$~\cite{SSFNet}& LDVAE-T \\ 
Endmember&2017&2022&2024&2024&2024&2025\\\midrule
Asphalt road    &-&-& 0.2889 & 0.1566 & 0.1578 & \textbf{0.03538}~\textcolor{OliveGreen}{\scriptsize$\uparrow$77.41\%} \\
Grass           &-&-& 0.1832 & 0.1977 & 0.1416 & \textbf{0.03083}~\textcolor{OliveGreen}{\scriptsize$\uparrow$78.23\%} \\
Tree            &-&-& 0.1737 & 0.1632 & 0.1179 & \textbf{0.02576}~\textcolor{OliveGreen}{\scriptsize$\uparrow$78.15\%} \\
Roof            &-&-& 0.1250 & 0.1283 & 0.0909 & \textbf{0.02476}~\textcolor{OliveGreen}{\scriptsize$\uparrow$72.76\%} \\
Metal           &-&-& 0.2599 & 0.0992 & -      & \textbf{0.03830}~\textcolor{OliveGreen}{\scriptsize$\uparrow$61.39\%} \\
Dirt            &-&-& 0.1334 & 0.1894 & 0.1192 & \textbf{0.03128}~\textcolor{OliveGreen}{\scriptsize$\uparrow$73.76\%} \\ \midrule
Average         &-&-& 0.1840 & 0.1558 & 0.1256 & \textbf{0.03105}~\textcolor{OliveGreen}{\scriptsize$\uparrow$75.28\%} \\ \bottomrule
\end{tabular}
\end{adjustbox}
}
\end{table*}

\section{Results}

We evaluate the proposed model on two key hyperspectral unmixing tasks: (1) abundance estimation and (2) endmember extraction.

\paragraph{Abundance Estimation.}
Tables~\ref{tab:RMSE_Samson},~\ref{tab:RMSE_Jasper}, and~\ref{tab:RMSE_Urban} report Root Mean Squared Error (RMSE) for abundance estimation on the Samson, Jasper Ridge, and HYDICE Urban datasets. Our model consistently achieves lower RMSE across all endmembers and datasets, outperforming state-of-the-art baselines. This performance is attributable to the Dirichlet prior imposed in the latent space, which naturally enforces the sum-to-one and non-negativity constraints expected of abundance vectors, yielding stable and physically meaningful estimates. By contrast, methods based on unconstrained regression can violate these constraints or require \textit{ad hoc} normalization.
Figures~\ref{fig:samson_heatmaps},~\ref{fig:jasper_heatmaps}, and~\ref{fig:hydice_heatmaps} show heatmaps of the predicted abundances vs. ground truth abundances for the Samson, Jasper Ridge, and HYDICE Urban datasets.

\paragraph{Endmember Extraction.}
Tables~\ref{tab:SAD_Samson},~\ref{tab:SAD_Jasper}, and~\ref{tab:SAD_Urban} present Spectral Angle Distance (SAD) for the Samson, Jasper Ridge, and HYDICE Urban datasets. Our model also outperforms state-of-the-art methods by a significant margin across all datasets and endmembers, achieving uniformly lower SAD. This improvement can be attributed to the model's use of endmembers bundles to more accurately estimate the distribution of endmembers that exist within a real world scene. Figures~\ref{fig:SamsonSpectra},~\ref{fig:JasperSpectra}, and~\ref{fig:HYDICESpectra} show extracted vs. ground truth spectra for Samson, Jasper Ridge, and HYDICE Urban datasets.

\section{Conclusion and Future Work}

We introduce the \emph{Latent Dirichlet Transformer Variational Autoencoder} (LDVAE-T), a framework for hyperspectral unmixing that couples transformer-based modeling with a Dirichlet latent structure. By imposing a Dirichlet prior in the latent space, the model naturally enforces the sum-to-one and non-negativity constraints required for physically plausible abundance estimates. Beyond this probabilistic scaffold, LDVAE-T contributes a decoder based on \emph{bundled endmembers}: instead of treating each material as a single, fixed prototype, the decoder predicts for each patch a distributional prototype comprising a mean spectrum and a structured covariance defined over spectral segments. Reconstructions are formed by mixing these bundles with Dirichlet-distributed abundances, enabling the network to capture intrinsic intra-class variability while preserving interpretability. Experiments on three standard benchmarks—\emph{Samson}, \emph{Jasper Ridge}, and \emph{HYDICE Urban}—show that LDVAE-T delivers state-of-the-art performance in both endmember extraction and abundance estimation, as measured by spectral angle distance and root mean squared error, respectively.

The empirical gains of LDVAE-T stem from the combining of three ingredients: 1) a \emph{physically meaningful latent space} via the Dirichlet prior, which allows for a VAE framework with a distribution that works within the constraints of a mixture; 2) \emph{distributional endmembers} that explicitly model spectral variability present in real world scenes; and 3) \emph{transformer-based encoding} that leverages long-range spectral–spatial dependencies often missed by MLP or CNN encoders. We find that Segment-wise covariance parameterization strikes a balance between flexibility (capturing correlated variation within bands) and tractability (avoiding full dense covariances over all wavelengths). Collectively, these elements improve both abundance estimation and the stability of extracted endmember signatures.

    \bibliographystyle{ieeetr}
    \bibliography{myBib}

\end{document}

%% file: arxivpreamble.tex
\usepackage{amsmath, amssymb, amsfonts, amsthm}
\usepackage[dvipsnames]{xcolor}
\usepackage{adjustbox}
\usepackage[numbers,sort&compress]{natbib}

\newcommand{\etal}{\textit{et al.}}

%% file: myBib.bib
@misc{JasperRidge,
title = {{AVIRIS} Jasper Ridge Hyperspectral Dataset},
author = {{NASA} Jet Propulsion Laboratory},
year = {1995},
howpublished = {\url{https://aviris.jpl.nasa.gov/}}, note = {Accessed: 2025-04-30}
}

@ARTICLE{NMFQMV,
  author={Zhao, Min and Gao, Tiande and Chen, Jie and Chen, Wei},
  journal={IEEE Geoscience and Remote Sensing Letters}, 
  title={Hyperspectral Unmixing via Nonnegative Matrix Factorization With Handcrafted and Learned Priors}, 
  year={2022},
  volume={19},
  number={},
  pages={1-5},
  doi={10.1109/LGRS.2020.3047481}}

@ARTICLE{CNNAEU,

  author={Palsson, Burkni and Ulfarsson, Magnus O. and Sveinsson, Johannes R.},

  journal={IEEE Transactions on Geoscience and Remote Sensing}, 

  title={Convolutional Autoencoder for Spectral–Spatial Hyperspectral Unmixing}, 

  year={2021},

  volume={59},

  number={1},

  pages={535-549},

  doi={10.1109/TGRS.2020.2992743}}

@ARTICLE{SSFNet,
  author={Wang, Bin and Yao, Huizheng and Song, Dongmei and Zhang, Jie and Gao, Han},
  journal={IEEE Journal of Selected Topics in Applied Earth Observations and Remote Sensing}, 
  title={SSF-Net: A Spatial–Spectral Features Integrated Autoencoder Network for Hyperspectral Unmixing}, 
  year={2024},
  volume={17},
  number={},
  pages={1781-1794},
  doi={10.1109/JSTARS.2023.3327549}}

@ARTICLE{DuanUnDat,
  author={Duan, Yuexin and Xu, Xia and Li, Tao and Pan, Bin and Shi, Zhenwei},
  journal={IEEE Transactions on Geoscience and Remote Sensing}, 
  title={UnDAT: Double-Aware Transformer for Hyperspectral Unmixing}, 
  year={2023},
  volume={61},
  number={},
  pages={1-12},
  doi={10.1109/TGRS.2023.3310155}}

@misc{16x16words,
author={Dosovitskiy,Alexey and Beyer,Lucas and Kolesnikov,Alexander and Weissenborn,Dirk and Zhai,Xiaohua and Unterthiner,Thomas and Dehghani,Mostafa and Minderer,Matthias and Heigold,Georg and Gelly,Sylvain and Uszkoreit,Jakob and Houlsby,Neil},
year={2021},
title={An Image is Worth 16x16 Words: Transformers for Image Recognition at Scale},
journal={arXiv.org},
note={Copyright - © 2021. This work is published under http://arxiv.org/licenses/nonexclusive-distrib/1.0/ (the “License”). Notwithstanding the ProQuest Terms and Conditions, you may use this content in accordance with the terms of the License; Last updated - 2024-10-16},
language={English},
url={http://search.proquest.com.uproxy.library.dc-uoit.ca/working-papers/image-is-worth-16x16-words-transformers/docview/2453831057/se-2},
}

@ARTICLE{GhoshTransformer,
  author={Ghosh, Preetam and Roy, Swalpa Kumar and Koirala, Bikram and Rasti, Behnood and Scheunders, Paul},
  journal={IEEE Transactions on Geoscience and Remote Sensing}, 
  title={Hyperspectral Unmixing Using Transformer Network}, 
  year={2022},
  volume={60},
  number={},
  pages={1-16},
  doi={10.1109/TGRS.2022.3196057}}

@ARTICLE{YangTransformer,
  author={Yang, Zhiru and Xu, Mingming and Liu, Shanwei and Sheng, Hui and Wan, Jianhua},
  journal={IEEE Transactions on Geoscience and Remote Sensing}, 
  title={UST-Net: A U-Shaped Transformer Network Using Shifted Windows for Hyperspectral Unmixing}, 
  year={2023},
  volume={61},
  number={},
  pages={1-15},
  doi={10.1109/TGRS.2023.3321839}}

@ARTICLE{kiran,
  author={Mantripragada, Kiran and Qureshi, Faisal Z.},
  journal={IEEE Transactions on Geoscience and Remote Sensing}, 
  title={Hyperspectral Pixel Unmixing With Latent Dirichlet Variational Autoencoder}, 
  year={2024},
  volume={62},
  number={},
  pages={1-12},
  doi={10.1109/TGRS.2024.3357589}}

@article{phy1,
author = {Sun, Lingzhi and Lucey, Paul G.},
title = {Unmixing Mineral Abundance and Mg\# With Radiative Transfer Theory: Modeling and Applications},
journal = {Journal of Geophysical Research: Planets},
volume = {126},
number = {2},
pages = {e2020JE006691},
doi = {https://doi.org/10.1029/2020JE006691},
url = {https://agupubs.onlinelibrary.wiley.com/doi/abs/10.1029/2020JE006691},
eprint = {https://agupubs.onlinelibrary.wiley.com/doi/pdf/10.1029/2020JE006691},
note = {e2020JE006691 2020JE006691},
year = {2021}
}

@misc{phy2,
author={Janiczek,John and Thaker,Parth and Dasarathy,Gautam and Edwards,Christopher S. and Christensen,Philip and Jayasuriya,Suren},
year={2020},
title={Differentiable Programming for Hyperspectral Unmixing using a Physics-based Dispersion Model},
journal={arXiv.org},
note={Copyright - © 2020. This work is published under http://arxiv.org/licenses/nonexclusive-distrib/1.0/ (the “License”). Notwithstanding the ProQuest Terms and Conditions, you may use this content in accordance with the terms of the License; Last updated - 2020-07-15},
language={English},
url={http://search.proquest.com.uproxy.library.dc-uoit.ca/working-papers/differentiable-programming-hyperspectral-unmixing/docview/2423679118/se-2},
}

@ARTICLE{phy3,
  author={Drumetz, Lucas and Chanussot, Jocelyn and Jutten, Christian},
  journal={IEEE Geoscience and Remote Sensing Letters}, 
  title={Spectral Unmixing: A Derivation of the Extended Linear Mixing Model From the Hapke Model}, 
  year={2020},
  volume={17},
  number={11},
  pages={1866-1870},
  doi={10.1109/LGRS.2019.2958203}}

@ARTICLE{nnmf1,
  author={Zhang, Shaoquan and Zhang, Guorong and Li, Fan and Deng, Chengzhi and Wang, Shengqian and Plaza, Antonio and Li, Jun},
  journal={IEEE Transactions on Geoscience and Remote Sensing}, 
  title={Spectral-Spatial Hyperspectral Unmixing Using Nonnegative Matrix Factorization}, 
  year={2022},
  volume={60},
  number={},
  pages={1-13},
  doi={10.1109/TGRS.2021.3074364}}

@ARTICLE{nnmf2,
  author={Wang, Xinyu and Zhong, Yanfei and Zhang, Liangpei and Xu, Yanyan},
  journal={IEEE Transactions on Geoscience and Remote Sensing}, 
  title={Spatial Group Sparsity Regularized Nonnegative Matrix Factorization for Hyperspectral Unmixing}, 
  year={2017},
  volume={55},
  number={11},
  pages={6287-6304},
  doi={10.1109/TGRS.2017.2724944}}

@ARTICLE{nnmf3,
  author={He, Wei and Zhang, Hongyan and Zhang, Liangpei},
  journal={IEEE Transactions on Geoscience and Remote Sensing}, 
  title={Total Variation Regularized Reweighted Sparse Nonnegative Matrix Factorization for Hyperspectral Unmixing}, 
  year={2017},
  volume={55},
  number={7},
  pages={3909-3921},
  doi={10.1109/TGRS.2017.2683719}}

@ARTICLE{nnmf4,
  author={Lu, Xiaoqiang and Wu, Hao and Yuan, Yuan and Yan, Pingkun and Li, Xuelong},
  journal={IEEE Transactions on Geoscience and Remote Sensing}, 
  title={Manifold Regularized Sparse NMF for Hyperspectral Unmixing}, 
  year={2013},
  volume={51},
  number={5},
  pages={2815-2826},
  doi={10.1109/TGRS.2012.2213825}}

@ARTICLE{DeepGUn,
  author={Borsoi, Ricardo Augusto and Imbiriba, Tales and Bermudez, José Carlos Moreira},
  journal={IEEE Transactions on Computational Imaging}, 
  title={Deep Generative Endmember Modeling: An Application to Unsupervised Spectral Unmixing}, 
  year={2020},
  volume={6},
  number={},
  pages={374-384},
  doi={10.1109/TCI.2019.2948726}}

@INPROCEEDINGS{soham,
  author={Chitnis, Soham and Mantripragada, Kiran and Qureshi, Faisal Z.},
  booktitle={IGARSS 2024 - 2024 IEEE International Geoscience and Remote Sensing Symposium}, 
  title={SpACNN-LDVAE: Spatial Attention Convolutional Latent Dirichlet Variational Autoencoder for Hyperspectral Pixel Unmixing}, 
  year={2024},
  volume={},
  number={},
  pages={7714-7719},
  doi={10.1109/IGARSS53475.2024.10640940}}

@misc{hydiceurban,
  title        = {HYDICE Urban Hyperspectral Dataset},
  author       = {U.S. Army Corps of Engineers},
  year         = {1995},
  note         = {Captured by the Hyperspectral Digital Imagery Collection Experiment (HYDICE) sensor. Available via academic requests.}
}

@misc{samsondataset,
  title        = {Samson Hyperspectral Dataset},
  author       = {Hyperspectral Imaging Laboratory},
  year         = {2011},
  note         = {Publicly available dataset commonly used for hyperspectral unmixing studies.}
}
